\documentclass{article}

\usepackage{microtype}
\usepackage{graphicx}
\usepackage{subfigure}
\usepackage{booktabs} 

\usepackage{amsmath}
\usepackage{amssymb}
\usepackage{mathtools}

\usepackage{rotating}
\usepackage{pdflscape}

\usepackage{hyperref}

\usepackage[accepted]{icml2021_arxiv}

\icmltitlerunning{Robust and integrative Bayesian neural networks for likelihood-free parameter inference}

\begin{document}

\twocolumn[
\icmltitle{Robust and integrative Bayesian neural networks\\ for likelihood-free parameter inference}


\icmlsetsymbol{equal}{*}

\begin{icmlauthorlist}
\icmlauthor{Fredrik Wrede}{equal,uu}
\icmlauthor{Robin Eriksson}{equal,uu}
\icmlauthor{Richard Jiang}{ucsb}
\icmlauthor{Linda Petzold}{ucsb}
\icmlauthor{Stefan Engblom}{uu}
\icmlauthor{Andreas Hellander}{uu}
\icmlauthor{Prashant Singh}{umu}
\end{icmlauthorlist}

\icmlaffiliation{uu}{Department of Information Technology, Uppsala University, Uppsala, Sweden}
\icmlaffiliation{ucsb}{Department of Computer Science, University of California-Santa Barbara, Santa Barbara, CA, USA}
\icmlaffiliation{umu}{Department of Computing Science, Umeå University, Umeå, Sweden}

\icmlcorrespondingauthor{Fredrik Wrede}{fredrik.wrede@it.uu.se}
\icmlcorrespondingauthor{Prashant Singh}{prs@cs.umu.se}


\icmlkeywords{Bayesian Approximate Inference, Active Learning, Supervised Learning, Computational Biology, Bayesian Deep Learning}

\vskip 0.3in
]



\printAffiliationsAndNotice{\icmlEqualContribution} 
\begin{abstract}
State-of-the-art neural network-based methods for learning summary statistics have delivered promising results for simulation-based likelihood-free parameter inference.
Existing approaches for learning summarizing networks are based on deterministic neural networks, and do not take network prediction uncertainty into account. 
This work proposes a robust integrated approach that learns summary statistics using Bayesian neural networks, and produces a proposal posterior density using categorical distributions. 
An adaptive sampling scheme selects simulation locations to efficiently and iteratively refine the predictive proposal posterior of the network conditioned on observations.
This allows for more efficient and robust convergence on comparatively large prior spaces. 
The approximated proposal posterior can then either be processed through a correction mechanism to arrive at the true posterior, or be used in conjunction with a density estimator to obtain the posterior.
We demonstrate our approach on benchmark examples.
\end{abstract}

\section{Introduction}
\label{sec:intro}
 Likelihood-free inference (LFI) is a class of methods designed for parameter inference of models with intractable or expensive likelihoods. 
 Approximations to the posterior distribution conditioned on observed data are made with the aid of forward simulations of the model, thus also referred to as simulation-based inference~\cite{Cranmer_frontierSBI}. Approximate Bayesian computation (ABC)~\cite{sisson2018handbook} is a popular method in this setting, and involves the use of a discrepancy measure to compare simulated data against observed data. A threshold on the discrepancy measure decides whether a proposed parameter can be accepted as a sample from the posterior. 
 
While this approach has shown to be robust, it has certain drawbacks. For many dynamical models, the simulation output is high-dimensional, 
and one uses summary statistics as a dimension reduction technique
 \cite{sisson2018handbook,wood2010statistical,engblom2020bayesian}. Selecting informative summary statistics so that the discrepancy measure can accurately accept posterior samples is often a tedious task, typically requiring detailed domain knowledge together with a trial and error approach. 
 
 An alternative approach to handpicking summary statistics is to utilize embedded neural networks (NN) to learn summary statistics from simulated data \cite{jiang2017learning, dinev2018dynamic, wiqvist2019partially, aakesson2020convolutional,chen2020neural}.  Some of these methods require a post-processing ABC scheme to yield the posterior, others work in conjunction with neural density estimators \cite{lueckmann2021benchmarking} to either estimate the posterior directly, or use synthesized likelihoods which requires Markov chain Monte Carlo (MCMC) sampling. Although deterministic NNs for learning summary statistics have shown impressive results, they are prone to be inaccurate in regions where the training data is sparse or not well represented (particularly in the case of problems with high-data dimensionality) \cite{aakesson2020convolutional}. 
Deterministic NNs do not directly provide an estimate of their prediction uncertainty, which can potentially mislead the parameter inference process.
This lack of direct uncertainty quantification motivates the design of more robust and expressive embedding networks, that can drive the LFI process in a more reliable manner.
 

Bayesian embedded NNs are proposed as summary statistics that provide prediction uncertainties, and coupled with a distribution estimator, also approximate the proposal posterior. An adaptive sampling algorithm utilizes these uncertainty estimates to iteratively refine the approximated proposal posterior. Including prediction uncertainty in the LFI process provides robustness against potential gaps in training data.

The true posterior distribution can be obtained from a correction procedure applied to the proposal posterior (out of scope of this work). The need for the correction arises due to the iterative nature of the adaptive sampling algorithm. The proposal posterior is refined in each iteration by sampling solely within the proposal posterior of the previous iteration. Therefore, in order to account for \emph{proposal leakage} or regions within the true posterior missed due to inaccuracies in the NN, a correction mechanism is needed that enlarges the proposal posterior accordingly.

The proposed approach involves approximating the proposal posterior through discrete categorical distributions, subsequently solving a regression problem via classification. This approach in similar form has delivered encouraging results~\cite{weiss1995rule, torgo1996regression, AHMAD_RvC}. The main motivation behind this design is that performing regression with high-dimensional inputs (such as dynamical simulation responses) to continuous multi-dimensional outputs (such as parameters) is a highly challenging task. Even though NNs have the learning capacity to do so, generalizing the networks often proves difficult. 

 We demonstrate the efficiency and robustness of our approach on benchmark problems, and also compare the proposal posteriors against approximated posteriors obtained using deterministic NNs paired with neural density estimators. Experimental results show  that the proposed approach is consistently more robust than existing methods, with relatively small variance in parameter inference quality performance. The error in parameter estimation even without the correction process is comparable to existing methods. 
We organize the paper as follows: Section~\ref{sec:background} introduces the problem setting, the notation, and related work. Section~\ref{sec:method} presents the proposed LFI parameter estimation approach. Section~\ref{sec:experiments} demonstrates the proposed methodology on benchmark examples and draws comparisons to other methods. Section~\ref{sec:discussion} presents a discussion on the merits and applicability of the approach. Finally, Section~\ref{sec:conclusion} gives concluding remarks about the novel method presented in the paper.

\section{Background} 
\label{sec:background}
Let ${\bf y}_0$ be a target observation which can be approximately modeled by $M({\bf y} | {\boldsymbol \theta})$.
Forward simulation of the model $M$ results in the sample path ${\bf y}^{1:T}$ henceforth referred to as ${\bf y}$.
The objective is to infer the parameters conditioned on the observation, i.e., the posterior density $p(\boldsymbol{\theta} | {\bf y}_0) \propto p({\bf y}_0 | \boldsymbol{\theta})p(\boldsymbol{\theta})$. However, as we assume that the likelihood is computationally intractable, which is often the case for dynamical models, an estimate $\hat{p}(\boldsymbol{\theta} | {\bf y}_0)$ of the true posterior distribution must be approximated. Approaches in this setting typically utilize access to the observed data  ${\bf y}_0$ and the ability to perform simulations from model $M$ at well-chosen parameter locations. We summarize some the methodologies under the supposed setting below.

\subsection{Related work}
The ABC family of methods forms the most popular and established LFI approach (introduced in Section~\ref{sec:intro}). 
Although remarkably successful, ABC methods are typically challenging to setup optimally for accurate inference and convergence speed \cite{sisson2018handbook,marin2012approximate}. Extensions such as sequential Monte Carlo methods (ABC-SMC) \cite{sisson2007sequential} can potentially accelerate the convergence speed, but still require laborious hyperparameter selection. In particular, the choice of informative summary statistics, is still present, and is critical for practical and accurate parameter inference as it forms the means of comparing simulated proposals with the observation. 

Consequently, summary statistic selection has received considerable attention \cite{sisson2018handbook}[Chapter 5]. However, traditional summary statistics such as statistical moments may not always be expressive enough to capture intricate details in rich dynamical data. Further, ideal statistics may even be missing from candidate pool of statistics to select from, e.g., \emph{tsfresh}~\cite{christ2018time}. A recent development has been the use of data-driven regression models to extract features from data, and function as summary statistics. It has been shown that the ideal summary statistic is the posterior mean, when minimizing the quadratic loss \cite{fearnhead2012constructing}. In particular, NN based regression models have delivered highly promising results towards learning the estimated posterior mean as summary statistics from high-dimensional detailed time series \cite{jiang2017learning,wiqvist2019partially,aakesson2020convolutional}. Such models learn the mapping $F_{\bf w}({\bf y}) \mapsto \mathbb{E}[\boldsymbol{\theta}]$ which can be seen as an amortized point estimate for a given ${\bf y}$, where ${\bf w}$ are the weights of the regression model. However, an overhead associated with \emph{trainable} summary statistics is the need for large amounts of simulated training data to yield an accurate  posterior mean estimate. Although there are opportunities for re-using the training data in the later rejection sampling step to estimate the conditioned posterior, the sheer amount of training data required is problematic whenever the simulations are expensive to compute. 


Neural density estimators form a family of methods that directly estimate densities with the aid of NNs and have shown impressive performance on challenging problems \cite{lueckmann2021benchmarking}, such as the masked autoregressive flows (MAF) and the neural spline flow (NSF) \cite{papamakarios2017masked,NSF}. While some neural density estimators are used to estimate posterior densities \cite{APT_2019,SNPE_A_Papamakarios, SNPE_B_lueckmann2017flexible}, others rely on constructing synthesized likelihoods or ratios which require MCMC sampling to extract the posterior \cite{papamakarios2019sequential, lueckmann2019likelihood,SNRE_a, SNRE_b_constrative}. The formed posteriors are amortized, i.e., any observation can be conditioned on the posterior. 

However, a collection of algorithms based on neural density estimators uses sequential Bayesian updates to the posterior, conditioned on a particular target observation. These include sequential neural posterior estimation (SNPE) \cite{SNPE_A_Papamakarios, SNPE_B_lueckmann2017flexible, APT_2019},  sequential neural likelihood estimation (SNLE) \cite{papamakarios2019sequential, lueckmann2019likelihood} and sequential neural ratio estimation (SNRE) \cite{SNRE_a, SNRE_b_constrative}. A benchmarking effort was recently conducted comparing the performance of these approaches on various problems \cite{lueckmann2021benchmarking}. These approaches also support the use of embedding networks to turn high-dimensional simulation outputs to informative summary statistics. The weights of the embedding networks are then trained together with the weights of the neural density estimator. Thus, these shared architectures enable sequential updates on the conditioned posterior and the summary statistics learning process, ultimately reducing the amount of simulations required. 

SNPE-C (also known as automatic posterior transform - APT) \cite{APT_2019} is a recent SNPE technique that learns to infer the true posterior by maximizing an intermediate estimated proposal posterior. SNPE approaches form a good class of methods to compare the proposed approach to, as the use of sequential Bayesian updates to refine the estimated posterior is close in spirit to adaptive sampling used in this work. We note that SNPE approaches approximate the true posterior, while the proposed approach needs a correction step as post-processing to arrive at the true posterior from a proposal posterior. SNPE approaches have also been shown to perform well across a variety of test problems \cite{lueckmann2021benchmarking,APT_2019}. Among SNPE approaches, SNPE-C is superior as it is more flexible and scalable as compared to SNPE-A and SNPE-B \cite{APT_2019}, and is used for the purpose of comparisons in the experiments in this paper.

    
%

\subsection{Motivation for Bayesian neural networks}
We implement a NN for summary statistics, with density estimation through Bayesian neural networks (BNNs). The introduction of weight uncertainties in NNs has enabled a robust means for learning an ensemble of models \cite{Bayesbybackprop}, represented by the posterior density $p({\bf w}|{\bf y})$ where ${\bf w}$ are the network weights.
It has also been demonstrated \cite{Bayesbybackprop} that the number of weights only increases by a factor of two by utilizing Monto Carlo (MC) sampling of gradients when updating the weights. The benefit of using a BNN is that the proposal posterior predictive distribution $\tilde{p}_w(\boldsymbol{\theta}|{\bf y}) = \int \tilde{p}(\boldsymbol{\theta}|{\bf w}, {\bf y})p({\bf w}|{\bf y})d{\bf w}$ provides an uncertainty measure to its predictions. The uncertainty estimates form an elegant means of identifying and countering overfitting, which translates into more robust and efficient embedding networks. Different embedding network architectures such as RNNs or CNNs can be implemented in the BNN setting to provide prediction uncertainty estimates, and compute the proposal posterior $\tilde{p}(\boldsymbol{\theta}|{\bf y}_0) \coloneqq \tilde{p}_w(\boldsymbol{\theta}|{\bf y}_0)$ by the proposal posterior predictive distribution. We will henceforth denote $\tilde{p}_w(\boldsymbol{\theta}|{\bf y})$ simply as $\tilde{p}(\boldsymbol{\theta}|{\bf y})$. The uncertainty estimates also allow sequential adaptive sampling of the posterior, conditioned on a target observation for faster convergence, similar to SNPE.

\section{Embedded Summary Statistics via Bayesian neural networks}
\label{sec:method}
 Let $D = \{({\bf y}_i,\boldsymbol{\theta}_i)\}_1^N$ consist of simulated data and parameter proposals sampled from the prior distribution over the parameter space $p(\boldsymbol{\theta})$. Consider a BNN, parameterized by the weights ${\bf w}$, that computes the posterior density $p({\bf w}|{\bf y})$ of a probabilistic model $\tilde{p}(\boldsymbol{\theta}|{\bf y},{\bf w})$ by placing a prior $p({\bf w})$ on the weights. The proposal posterior predictive distribution for new unseen sample path ${\bf y}_t$,  $\tilde{p}(\hat{\boldsymbol{\theta}}|{\bf y}_t) = \mathbb{E}_{p({\bf w}|D)}[\tilde{p}(\hat{\boldsymbol{\theta}}|{\bf y}_t, {\bf w})]$ is the expectation on unseen data. Setting ${\bf y}_t = {\bf y}_0$, for $\hat{\boldsymbol{\theta}} \approx \boldsymbol{\theta}$ the proposal posterior predictive distribution will be equal to $\tilde{p}(\boldsymbol{\theta}|{\bf y}_0)$ as $N \rightarrow \infty$, which is the target parameter proposal posterior conditioned on a particular observation. 
 As we demonstrate next, we will approximate the parameters $\boldsymbol{\theta}$ through discrete categories making ${\tilde p}(\boldsymbol{\theta}|{\bf y},{\bf w})$ a categorical distribution thereby transforming the regression problem into a classification problem. 

\subsection{Probabilistic regression via classification}
Let $\tilde{p}(\boldsymbol{\theta}'|{\bf y},{\bf w})$ be represented as a categorical distribution using cross-entropy and softmax activation in a NN classifier denoted by $F_{\tilde{p}(\boldsymbol{\theta}'|{\bf y},{\bf w})}$.  We will transform the continuous problem by discretizing $\boldsymbol{\theta} \in \mathbb{R}$ into $\boldsymbol{\theta}'\in K$ by $B$ equal width bins per dimension (please see Section~\ref{sec:discussion} for a discussion on binning strategies), i.e., the predictive distribution will provide the probability of belonging to a particular bin $K \in {1,2,...,B}$. By placing a multivariate Gaussian on $p({\bf w})$ we will fit $p({\bf w}|D)$ using Bayes by backprop, subsequently yielding the BNN classifier  $F_{p({\bf w}|D)}$. To estimate the proposal posterior predictive distribution conditioned on our target observation, we sample from $F_{p({\bf w}|D)}$ using a MC sampling scheme,   
\begin{equation}
    \tilde{p}(\boldsymbol{\theta}'|{\bf y}_0) \approx \frac{1}{N}\sum_{i=1}^N F_{p({\bf w}|D)}({\bf y}_0, {\bf w}_i).
\end{equation}

\begin{figure}[t]
\vskip 0.2in
\begin{center}
\centerline{\includegraphics[width=\columnwidth]{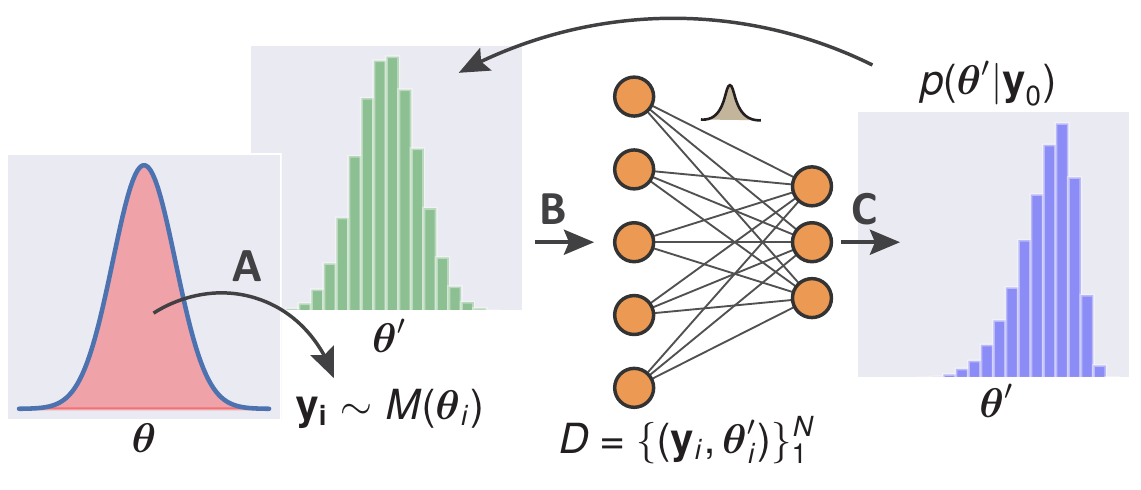}}
\caption{Adaptive sampling-driven BNNs. (A) Parameter proposals are sampled from a prior and simulated via the model. (B) Regression via classification. Continuous proposals are discretized into a categorical distribution with classes being the bins. A BNN (having prior weights) is then trained on data. (C) The posterior predictive conditioned on observation $y_0$ can be estimated via MC sampling, and is subsequently used in an iterative procedure, replacing the previous prior. Note that a transformation is applied to the categorical distribution to transform proposals back to continuous.}
\label{fig:pipeline}
\end{center}
\vskip 0.2in
\end{figure}

To enable sequential adaptive sampling during the inference task, we will train $G$ subsequent classifiers $F^g_{p({\bf w}|D)}, \, g = 1, \dots, G$, where the bins are updated by sampling from the proposal posterior predictive distribution. Each iteration employs the density from the previous generation as the prior for sampling new training data. The BNN generates amortized proposal posterior estimates. However, when conditioned on the observation, it allows for adaptive refinement in each iteration. To be able to simulate new samples, we need to transform $\boldsymbol{\theta}'$ back to $\boldsymbol{\theta}$. This is achieved by transforming each bin $K$ into a uniform distribution,

\begin{equation}
\begin{aligned}
    \{\boldsymbol{\theta}'\}_{i=1}^N &\sim \tilde{p}(\boldsymbol{\theta}'|{\bf y}_0) \\
    \{\boldsymbol{\theta}\}_{i=1}^{N} &= \{\mathbb{U}^K({\boldsymbol{\theta}'}_i^{\text{\emph{left}}},{\boldsymbol{\theta}'}_i^{\text{\emph{right}}})\}_{i=1}^N,
\end{aligned}
\end{equation}

where \emph{left} and \emph{right} are the boundaries of the bin intervals. The adaptive sampling-driven BNN is depicted in Figure~\ref{fig:pipeline} and detailed in Algorithm~\ref{alg:posterior}.

\subsection{Exploitation of bins}
We aim to exploit the uncertainty estimates provided by the proposal posterior predictive distribution to enable faster convergence. A key reason why the process can be slow is that low probability weights linger between iteration updates. We exploit the proposal posterior predictive distribution by using a threshold value, $\delta$. Given the categorical distribution, we set the classes with probability weight below the threshold to zero. The non-zero probabilities are then normalized, constructing a less skewed proposal distribution. Between iterations, $\delta_i$ can be chosen adaptively or kept constant throughout. We recognize that the sequence of distinct $\delta_i$ holds a similarity to the discrepancy thresholds ($\epsilon_i$) used by ABC-SMC~\cite{sisson2007sequential, del2012adaptive}.
The potential adaptive nature is briefly discussed in Section~\ref{sec:discussion}.

\subsection{Dealing with the curse of dimensionality}
For the single-parameter problem, the structure of the classification problem is straightforward: one classifier for one parameter. However, for multiple parameter dimensionality, we construct one classifier for each two-dimensional covariance combination: $(\boldsymbol{\theta}_i, \boldsymbol{\theta}_j)$, implying that for a model of $k$ parameters, we would require $\frac{k!}{(k-2)!2!}$ classifiers. One sees that the number of classifiers quickly increases; however, they lend themselves towards an embarrassingly parallel workflow since they are independent, and use the same simulation data. Further, the adaptation of the proposals is done per covariance dimension and then marginalized, resulting in the $k$-dimensional parameter proposals.

\begin{algorithm}[]
   \caption{Iterative BNN proposal posterior estimation }
   \label{alg:posterior}
\begin{algorithmic}
   \STATE {\bfseries Input:} model simulator $M({\bf y}|\boldsymbol{\theta})$, 
   observations ${\bf y}_0$, Bayesian neural network $F_{p({\bf w}|D)}$, 
   number of generations $G$, number of simulations per generation $N$,
   prior distribution $p(\boldsymbol{\theta})$, number of bins $B$.

   \STATE $p_0 := p$  
   \FOR{$g=1$ {\bfseries to} $G$}
            \STATE Sample $\boldsymbol{\theta}_N = \{\boldsymbol{\theta}_i\}_{i=1}^N$ from $p_{g-1}(\boldsymbol{\theta})$
            \STATE Sample ${\bf y}_N = \{{\bf y}_i\}_{i=1}^N$ from $M({\bf y}|\boldsymbol{\theta}_i)$
        \STATE Group $\boldsymbol{\theta}_N$ into K bins $\boldsymbol{\theta}'_N, \, K= 1:B$
        \STATE Train $F^g_{p({\bf w}|D_g)}$ on $D_g = \{{\bf y}_N, \boldsymbol{\theta}'_N\}$
        \STATE Estimate $\tilde{p}_{g}(\boldsymbol{\theta}'_N|{\bf y}_0) \leftarrow \mathbb{E}[F^g_{p({\bf w}|D_g)}({\bf y}_0,{\bf w}_N)]$
        \STATE Transform $p_g(\boldsymbol{\theta}) \leftarrow \mathbb{U}^K({\tilde{p}}_{g}(\boldsymbol{\theta}'_N|{\bf y}_0))$
    \ENDFOR
    \STATE {\bfseries return} $p_g(\boldsymbol{\theta})$
\end{algorithmic}
\end{algorithm}




\section{Experiments}
\label{sec:experiments}

To test our approach we have implemented a convolutional BNN (we denoted this particular implementation BCNN). However, we note that our approach is not restricted to these types of architectures.  Our motivation behind choosing a convolutional architecture is that they have demonstrated excellent performance in identifying intricate patterns in complex, high-dimensional time-series data \cite{dinev2018dynamic, sadouk2019cnn, aakesson2020convolutional} as well as images~\cite{APT_2019}. For further information on the implementation we refer the reader to supplementary materials.

We explore the performance of the BCNN on two benchmark examples - the moving averages model (MA) and the discrete-state continuous-time Markov chain implementation of the Lotka-Volterra (LV) \cite{lotka}. Together with the resulting proposal posterior we recover using the BCNN, we also compare the proposal posterior estimation quality to posterior estimates obtained using ABC-SMC and SNPE-C with the MAF as the neural density estimator.

The SBI package \cite{tejero2020sbi} implements SNPE-C, and is used for the experiments in this paper. To keep the comparison fair between SNPE-C and the BCNN, we designed the embedded nets to be as similar as possible between the two approaches, resulting in having approximately the same number of weights in the two models.  
 
All of the experiments were conducted on a machine consisting of a 12-core Intel Xeon E5-1650 v3 processor operating at 3.5 GHz, 64 GB RAM, running Fedora core 32 Linux operating system. Even though GPU-acceleration would speed up NN training substantially, in the considered setting we assume that simulations are significantly more expensive than training time. This is the case for the discrete and stochastic LV model, less so for MA.

\subsection{Moving average model}
The moving average (MA) model is a popular benchmark problem in the ABC setting \cite{marin2012approximate}. The MA model of order $q$, MA($q$) is defined for observations $X_1, \dots, X_p$ as \cite{wiqvist2019partially,aakesson2020convolutional},
\begin{equation}
    X_j = Z_j + \theta_1 Z_{j-1} + \theta_2 Z_{j-2} + \dots + \theta_q Z_{j-q},
\end{equation}
where $j=1,\dots,p$, and $Z_j$ represents latent white noise. For the experiments in this paper, we consider $q=2$ and $Z_j \sim N(0,1)$ \cite{wiqvist2019partially}. The MA(2) model is identifiable in the following triangular region,
\begin{equation}
    \theta_1 \in [-2,2], \theta_2 \in [-1, 1], \theta_2 \pm \theta_1 \geq -1.
\end{equation}
In order to ease implementation and compatibility between the compared approaches, a rectangular uniform prior encompassing the triangular region was used in the experiments. We note here that the estimated (proposal) posterior was always within the triangular region, therefore no identifiability concerns were encountered due to a rectangular prior.


\subsubsection{Experimental Setup}
In the experiment we conduct a total of $G=6$ iterations (rounds) of refining the (proposal) posterior estimate for 10 individual seeds. The total amount of parameter proposals in each round is set to $N=3000$ and the number of bins for BCNN is set to $B=4$ per parameter dimension, resulting in a total of 16 bins. The exploitation threshold is set to $\delta = 0.05$. We simulate the MA(2) model with 100 equidistant time points.
For the ABC-SMC method, we use auto-correlation coefficients of lag 1 and 2, and variance as summary statistics. They are non-sufficient, but are chosen as a representation of a typical choice made by a practitioner in absence of much prior knowledge. Euclidean distance was used as the discrepancy measure.

\subsubsection{Results}
Here we demonstrate the perfomance of the proposed BCNN approach and compare against SNPE-C and ABC-SMC. The model has a superficial posterior, but the map $\boldsymbol{\theta} \mapsto {\bf y}$ is ill-discriminated, i.e., considerable variation in $\boldsymbol \theta$ provides little feedback in ${\bf y}$.  

\begin{figure}[ht]
\vskip 0.2in
\begin{center}
\centerline{\includegraphics[width=\columnwidth]{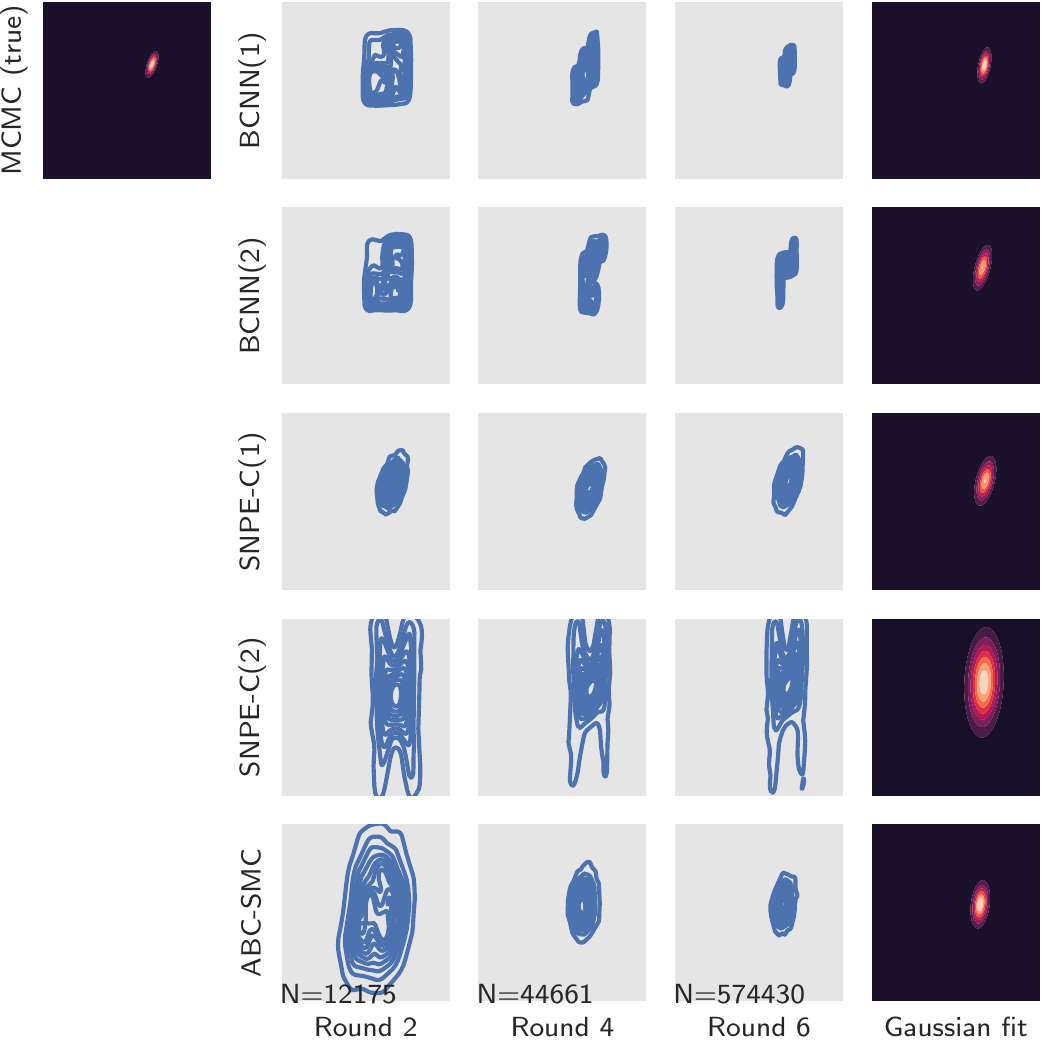}}
\caption{MA(2) per round proposal posterior estimation for BCNN, compared to approximated posteriors obtained using SNPE-C, and ABC-SMC, together with the ground truth - a MCMC generated posterior. Each round includes $N=3000$ parameter proposals, except for ABC-SMC which require many more. BCNN and SNPE-C includes two different random seeds (1) and (2). Final column to the right shows a Gaussian kernel density estimate of the final round posterior per method.}
\label{fig:grid_MA2}
\end{center}
\vskip 0.2in
\end{figure}

\begin{figure}[ht]

\includegraphics[width=\columnwidth]{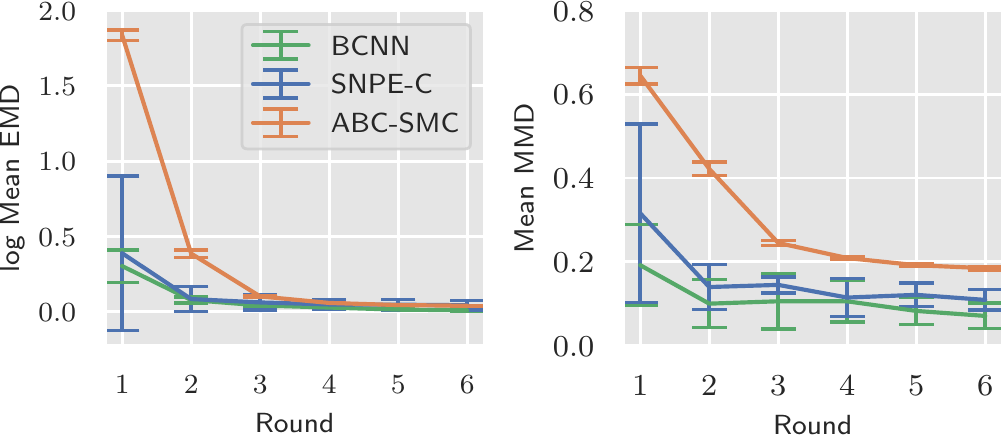}
\caption{Target posterior error estimates, log mean EMD (left) and Mean MMD (right) for BCNN, SNPE-C and ABC-SMC per round, with variability $\pm$ 1 std, recovered from ten independent seeds. The BCNN have a slight edge over SNPE-C, especially when it comes to the variability. Note that on average ABC-SMC will require more simulations to converge than the other two, see bottom row in Figure~\ref{fig:grid_MA2}.}
\label{fig:emd_MA2}

\end{figure}

Figure~\ref{fig:grid_MA2} provides qualitative insights on the shapes of the estimated proposal posterior obtained using the proposed approach, and posterior distributions obtained using the other approaches. Figure~\ref{fig:emd_MA2} depicts the log mean earth mover distance (EMD)~\cite{levina2001earth} and the mean of maximum mean discrepancy (MMD) \cite{Scholkopf_MMD} between the estimated distributions and the exact posterior. We see similar convergence patterns between BCNN and SNPE-C for both the EMD and MMD over subsequent rounds and different seeds, with the BCNN having a slight edge. This is despite the BCNN targeting a proposal posterior rather than the true posterior with a correction step. It can also be observed that the BCNN is highly robust over different seeds with very little variance in performance.
Another interesting observation from Figure~\ref{fig:grid_MA2} is the improvement or refinement extracted by each approach over subsequent rounds. It can be seen that BCNN(1) and BCNN(2) are able to substantially improve the quality of the estimated distributions between rounds 2-4 and 4-6. The ability of the BCNN architecture to greatly improve over successive rounds can be attributed to the combination of uncertainty estimates from the BCNN model leveraged by the adaptive sampling algorithm to identify regions most likely to contain the true posterior.


\subsection{Lotka-Volterra predator-prey model}
The Lotka-Volterra predator-prey dynamics model is a popular parameter estimation benchmark problem \cite{aakesson2020convolutional, greenberg2019automatic}. The model is implemented as a stochastic Markov jump process \cite{prangle2017adapting} simulated using the stochastic simulation algorithm (SSA) \cite{gillespie1977exact}. The model consists of three events - prey reproduction, predation and predator death. The model is described by the events:
\begin{equation}
 \begin{aligned}
     \mathcal{X}_1 &\xrightarrow{\theta_1} 2\mathcal{X}_1,\\
     \mathcal{X}_1 + \mathcal{X}_2 &\xrightarrow{\theta_2} 2\mathcal{X}_2,\\
     \mathcal{X}_2 &\xrightarrow{\theta_3} \emptyset.
 \end{aligned}
\end{equation}
The parameters ${\boldsymbol{\theta}} = \{ \theta_1, \theta_2, \theta_3\}$ control the events described above, and have corresponding rates $\theta_1 \mathcal{X}_1, \theta_2 \mathcal{X}_1 \mathcal{X}_2, \theta_3 \mathcal{X}_2$ respectively. 
\subsubsection{Experimental Setup}
We study the inference using a total of $G=8$ rounds of refining the target distribution, with five repetitions using different random number seeds, setting the total amount of parameter proposals in each round to  $N=1000$. Further, we define the number of bins to $B=5$ per dimension and the threshold to be $\delta = 0.05$.
The LV model data consists of 51 observations in the time span $[0 \text{ h},50 \text{ h}]$. We construct the model and the corresponding simulator using \textit{GillesPy2}\footnote{https://github.com/StochSS/GillesPy2} \cite{gillespy_2017}. 
Finally, the initial conditions include $\mathcal{X}_1=50$ and $\mathcal{X}_2=100$, and the actual parameters to be inferred are $[\log(1.0), \log(0.005), \log(1.0)]$. The prior we apply per dimension is $\mathbb{U}(\log(0.002), \log(2))$. We include the ABC-SMC estimate, because of the lack of a true posterior solution, and the LV model is a common benchmark problem in the ABC literature, i.e., the solution it produces is well accepted \cite{SNPE_A_Papamakarios,owen2015scalable}.
For the ABC-SMC method, we use no summary statistics, i.e., the data itself is directly compared, and a Euclidean distance as the discrepancy measure. The tolerance thresholds ($\epsilon$'s) for SMC-ABC are selected using a relative scheme, where the $20$-th percentile value of the ABC distances from the previous round is selected as $\epsilon$.
\subsubsection{Results}
\begin{figure}[ht]
\vskip 0.2in
\begin{center}
\centerline{\includegraphics[width=\columnwidth]{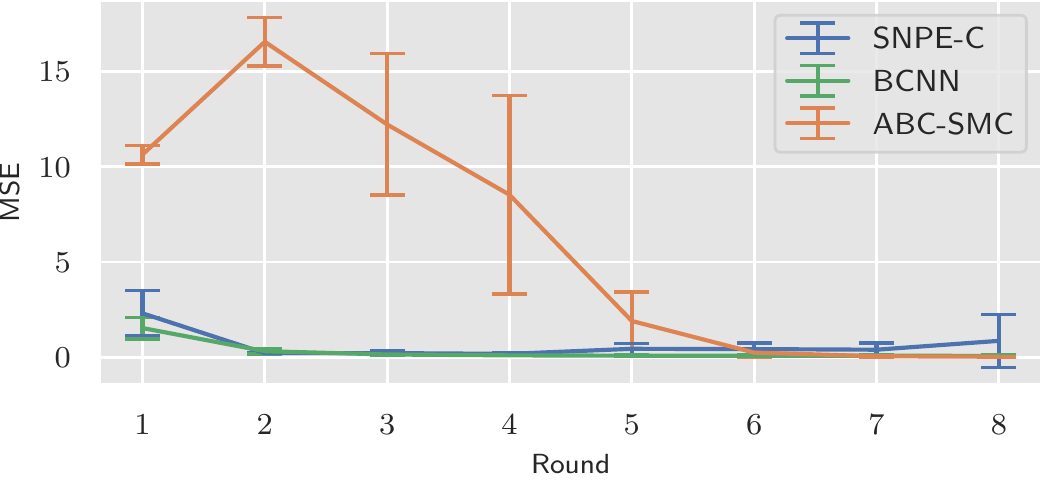}}
\caption{The lines with error bars show the mean MSE with variability $\pm1$ std, recovered from five seeds with eight rounds for SNPE-C, BCNN, and ABC-SMC. All methods converge to similar MSE at the final round, with low variability. Notable is the initial increase in MSE for ABC-SMC; from Figure~\ref{fig:grid_LV} it can be inferred that the approximated posterior (proposal posterior in case of BCNN) is bi-modal before finding a suitable discrepancy threshold.}
\label{fig:MSE_LV}
\end{center}
\vskip 0.2in
\end{figure}

\begin{figure}[ht]
\vskip 0.2in
\begin{center}
\centerline{\includegraphics[width=\columnwidth]{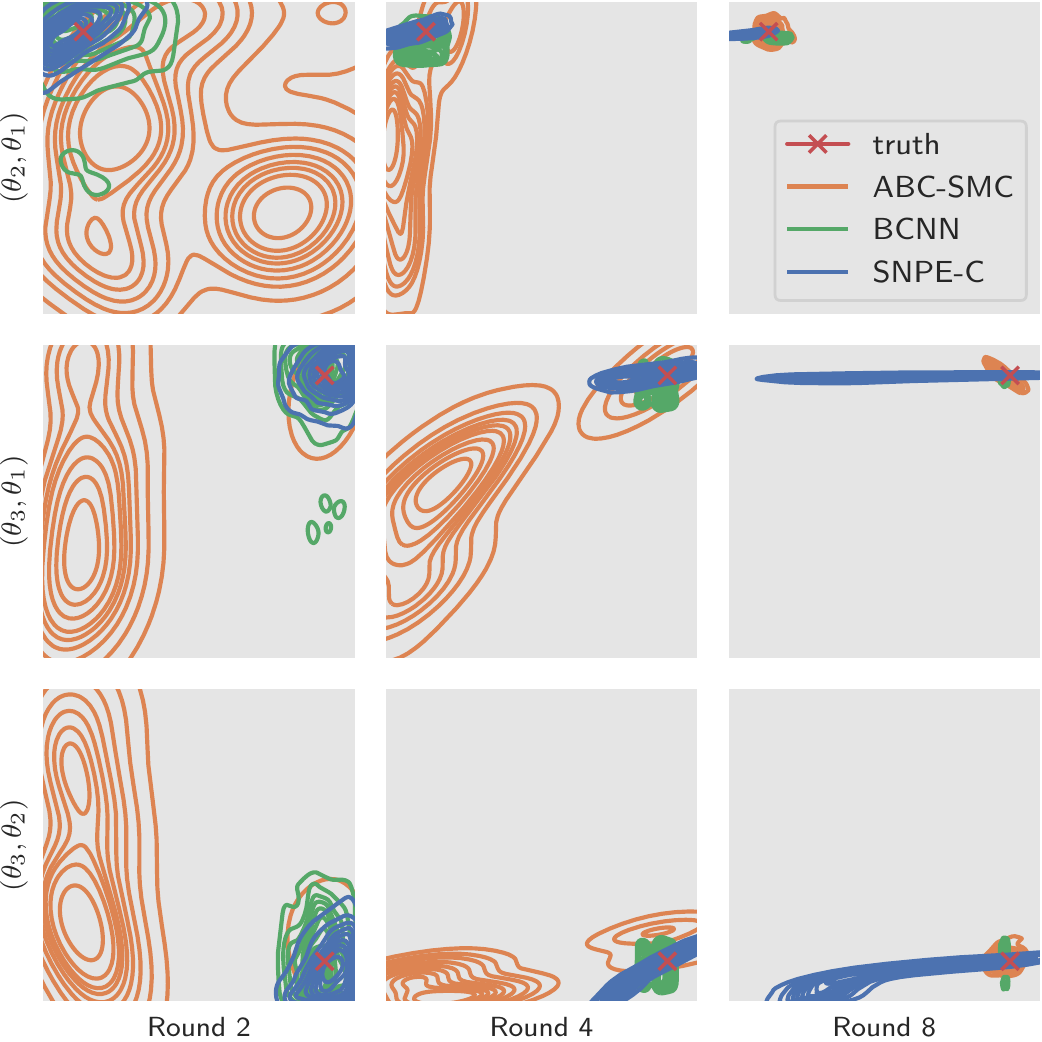}}
\caption{Snapshots of the proposal posterior estimate for the LV problem using BCNN, and posterior estimates using SNPE-C, and ABC-SMC. Each column is a different round: 2, 4, and the final 8, and each row gives the three parameters combinations possible of $\theta_1$, $\theta_2$, and $\theta_3$. The true parameter value is indicated by the orange cross. The prior covers the entire box uniformly. For the ABC-SMC the cumulative number of trials at round 2, 4 and 8 are $[8000, 19000, 127000]$, compared to $[2000, 4000, 8000]$ for BCNN and SNPE-C.}
\label{fig:grid_LV}
\end{center}
\vskip 0.2in
\end{figure}

Compared to the MA(2) model, we do not have an analytical posterior to compare to, thus we cannot refer to the EMD of the method. However, we solve the problem with known truth (the parameters to infer) and consider the mean square error (MSE) for computed target posteriors. The reasoning is that MSE holds information both about the bias and the variance of the distribution. But with the known limitation of not considering any higher moments. In Figure~\ref{fig:MSE_LV}, we evaluate the said MSE of the BCNN approach together with SNPE-C and ABC-SMC. The BCNN performs on par with SNPE-C, without any loss of posterior concentration; see post round 5 how the standard deviation increases per generation of SNPE-C. Exploring what might have caused the increase in standard deviation, we refer to Figure~\ref{fig:grid_LV}, which presents snapshots of a sequence of rounds from one run with the different methods. SNPE-C encounters an instability after round 4. From the MSE, we note that SNPE-C is mostly, not always, stable compared to the empirically robust BCNN.
Finally, we recognize that the initially poor performance of the ABC-SMC method can be attributed to the lack of adequately tuned summary statistics.

\section{Discussion}
\label{sec:discussion}

The results presented in the previous section demonstrate that the proposed approach is comparable to SNPE-C, which is a well-established method for approximating flexible posterior densities using NNs. Notably, the comparable performance has been achieved in spite of a lack of correction of the proposal posterior in the case of the BNN. The uncertainty estimates and a simplification strategy of accomplishing regression via classification using the BNN allows the adaptive sampling algorithm to consistently drive the search for the true posterior in the right direction. This can be seen in the very low standard deviations presented in Figures~\ref{fig:emd_MA2} and \ref{fig:MSE_LV}. 
However, an important observation is that SNPE-C does not utilize any exploitation similar to $\delta$ in the proposed approach  (see Section~\ref{sec:method}). Effectively, the deviations in performance are reduced using this exploitation. 

However, incorporating a similar exploitation strategy used in SNPE methods may not alone resolve the issue of occasional, substantially inaccurate posterior estimates, such as SNPE-C(2) in Figure \ref{fig:grid_MA2}. The stability and robustness of the proposed approach stems from the Bayesian setting considered in this work. Further, the BNN is specifically tailored for models with complex data distributions, such as discrete stochastic and dynamical models. 
\subsection{Future work}

The convergence plots in Figure~\ref{fig:emd_MA2} and \ref{fig:MSE_LV} point towards an opportunity  for early stopping when the proposal posterior estimate no longer receives significant updates. A possibility could be to measure the highest density region (HDR)~\cite{HDR}, or EMD, change between proposal posterior estimates, and stop if the difference is below a threshold value. Using this could substantially reduce the number of simulations involved and remove the requirement of specifying the amount of rounds $G$.

Choosing the optimal number of bins $B$ and number of proposals $N$ in each bin (support) is currently not trivial. A sufficient support for dynamical models is highly dependent on the response of simulation outputs to perturbations in the parameter space. For the MA(2) model, a larger $N$ per round was required, as well as fewer bins compared to the settings for the LV problem. This was owing to the MA(2) model mapping being ill-discriminated in nature as described earlier. Thus, for the BNN classifier to sufficiently learn discriminating features, one would require more data. Further, currently the bin width is fixed and does not consider the distribution of proposals. Exploring variable bin widths based on the rate of change of proposal distributions over successive rounds could potentially improve convergence speed, and also balance the support for bins more evenly. As a remark on the choice of number of bins, in \ref{sec:hyperMA2} and \ref{sec:hyperLV} we present the result from our empirical tests, which encourages further studies.

The procedure of assigning a continuous $\theta_i$ into a bin does not take discretization uncertainty into account (e.g., when $\theta$ is close to the egde of two bins), which can lead to inaccurate training. The aforementioned problems will be considered as future work and will involve adaptive bin selection based on the discriminative ability of the BNN. Further, bin uncertainties could potentially be tackled using fuzzy set theory and other bin strategies used in regression via classification.

The hyperparameter $\delta$ (see Section~\ref{sec:method}) that controls exploitation is currently considered to be fixed. However, an adaptive $\delta$ dependent on the current round can potentially improve convergence, e.g., an exponentially decaying $\delta$. The similarity to the threshold one use in ABC-SMC, aligns the adaptivity of $\delta$ well with strategies already in practice. In \ref{sec:hyperMA2} and \ref{sec:hyperLV} we give additional test results on the subject. 

A regression-based approach as opposed to the classification based approach presented in this paper will be explored. We will also work towards integrating the proposal correction process as part of the algorithm in order to estimate the true posterior.



\section{Conclusion}
\label{sec:conclusion}
This paper presented a novel Bayesian embedded neural network based approach for learning summary statistics for parameter inference of dynamical simulation models. The proposed approach leverages Bayesian neural networks for learning high-quality summary statistics as the estimated proposal posterior while also incorporating epistemic uncertainties. An adaptive sampling algorithm utilizes the estimated proposal posterior and uncertainty estimates from the Bayesian neural network to iteratively shrink the prior search space for the actual posterior distribution. An efficient classification-based density estimation approach further approximates the proposal posterior. The proposal posteriors obtained using the proposed approach are compared to approximated posteriors obtained using a well-established sequential neural posterior estimation method and the popular approximation Bayesian computation sequential Monte Carlo likelihood-free inference methods. Experiments performed on benchmark test problems show that the proposed approach is consistently more robust under various situations compared to existing methods while delivering comparable parameter inference quality.

\bibliography{references}
\bibliographystyle{icml2021}

\begin{appendix}
\onecolumn
\icmltitle{Robust and integrative Bayesian neural networks\\ for likelihood-free parameter inference\\
Supplementary Material}

\section{Extended Experiments for the MA(2) Test Problem:}
We investigate the effect of varying random seeds, and a comparison of execution times for the proposed approach against SNPE-C. We also explore the effect of varying the hyperparameters involved in the proposed BCNN-based approach.
\subsection{Qualitative evaluation with additional seeds}

\begin{figure*}[ht]
\begin{center}
\centerline{\includegraphics[width=\textwidth]{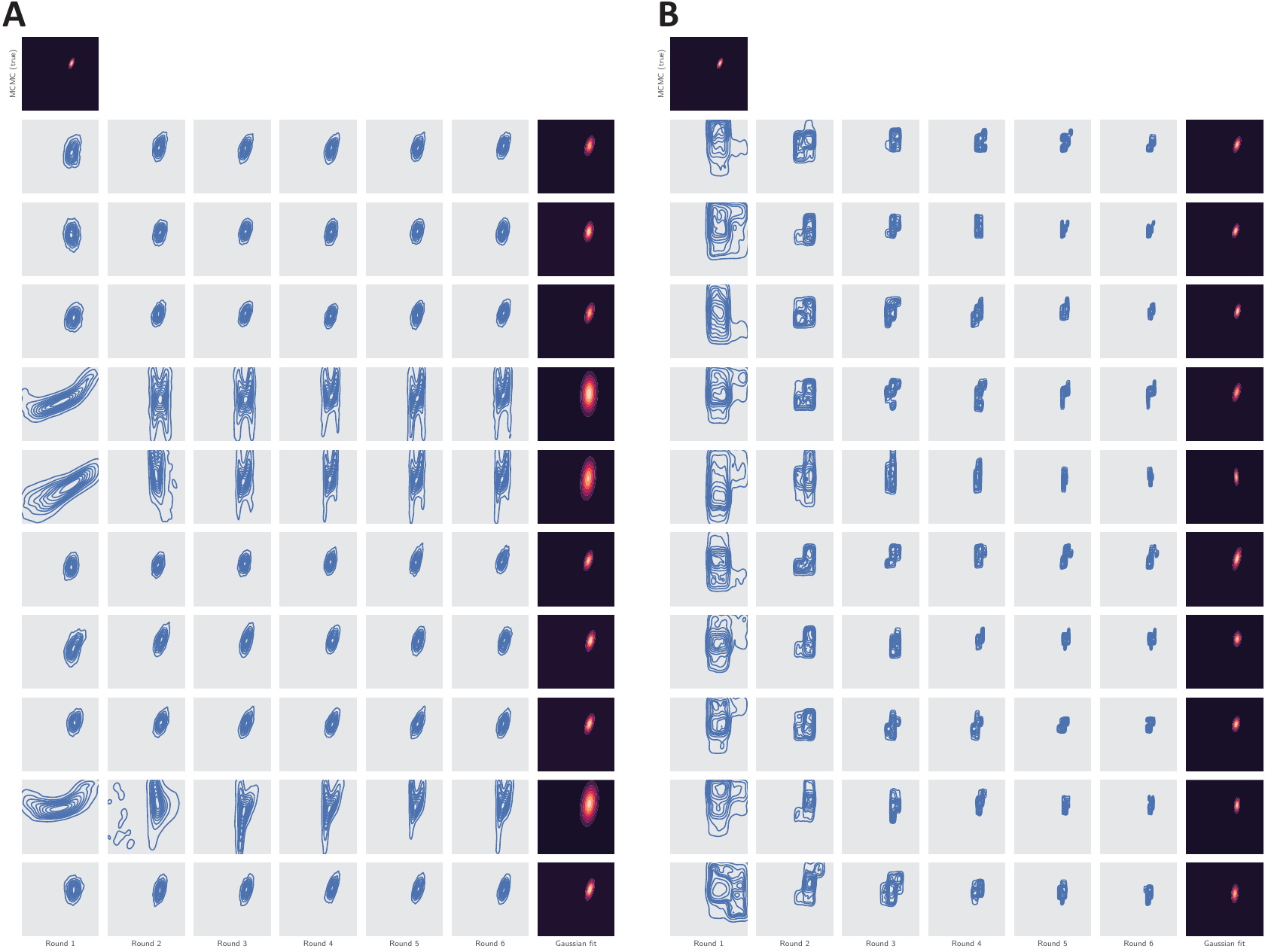}}
\caption{Ten individual seeds for estimating posterior using SNPE-C (A) and proposal posteriors using BCNN (B). }
\label{fig:ma2_grid_suppl}
\end{center}
\end{figure*}

\twocolumn
In order to investigate the robustness of the proposed approach, ten distinct random seeds were used in the parameter inference task for the BNN (BCNN) and SNPE-C based methods. Figure~\ref{fig:ma2_grid_suppl} depicts the results with each row representing a seed, and each column representing a round in the sequential parameter inference process. The random walk Metropolis-Hastings MCMC method was used to calculate the exact posterior for the purpose of comparisons with BCNN and SPNE-C. 

It can be observed that the proposed BCNN-based approach is consistent across different seeds, while SNPE-C shows susceptibility in three cases (seeds 4, 5 and 9). It can also be seen that SNPE-C typically starts with a tighter estimated posterior in the first round, but relative gains over subsequent rounds are small. In comparison, BCNN is slower in the initial stages, with larger estimated proposal posteriors in the first round. However, relative gains over successive rounds are more significant with substantial refinement being achieved. This can be attributed to the exploitation offered by the binning-based classification approach involved in parameter inference using BCNN. The final distribution in the sixth round is more concentrated for the BCNN as compared to SNPE-C. 

\subsection{Embedding Network architectures}
\label{sec:arch_ma2}
The BCNN was implemented using the TensorFlow library (Keras and tensorflow probability), while the SBI package (SNPE-C) makes use of PyTorch. The architectures are shown in Table \ref{tab:ma2}.
\begin{table}[ht]
\caption{Architecture for the BCNN (top) and SNPE-C (bottom); NDE refers to neural density estimator.}
\label{tab:ma2}
\begin{center}
\begin{tabular}{lcccr}
\toprule
Layer & Ouput shape & Weights\\
\midrule
Input    &[(None, 100, 1)]&0\\
Conv1DFlipout (ReLU) & (None, 100, 6)  &66\\
MaxPooling1D    & (None, 10, 6)&0\\
Flatten    & (None, 60)&0\\
DenseFlipout (Softmax)     &(None, 16)&1936\\
\midrule
Total weights &&2002

\end{tabular}

\begin{tabular}{lcccr}
\toprule
Layer & Ouput shape & Weights\\
\midrule
Input    &[(None, 100, 1)]&0\\
Conv1d (ReLU) &(None, 100, 6)  &36\\
MaxPooling1d    &(None, 10, 6)&0\\
view (Flatten)    & (None, 60)&0\\
Linear (ReLU)     &(None, 16)&488\\
NDE (MAF) &10 hidden, 2 transform&970\\
\midrule
Total weights &&1494

\end{tabular}
\end{center}
\end{table}

\subsection{Hyperparameters of the BNN}
\label{sec:hyperMA2}
\begin{figure}[ht]
\begin{center}
\centerline{\includegraphics[width=\columnwidth]{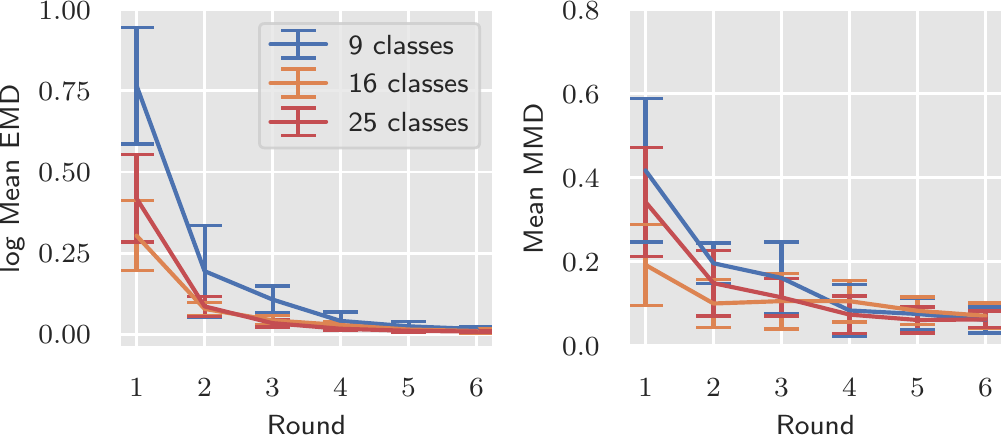}}
\caption{Log mean EMD and mean MMD for BCNN with different number of classes corresponding to the number of bins.}
\label{fig:ma2_bins}
\end{center}
\end{figure}

\begin{figure}[ht]
\begin{center}
\centerline{\includegraphics[width=\columnwidth]{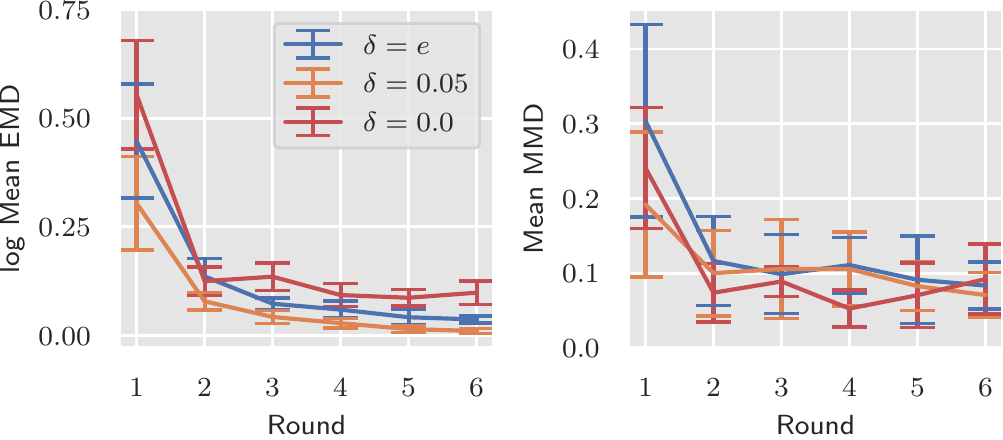}}
\caption{Log mean EMD and mean MMD for BCNN for different threshold $\delta$. $\delta = 0.0$ refer to no threshold being used (i.e no exploitation). $\delta = 0.05$ is the strategy used in the main article. $\delta = e$ is an exponentially decaying threshold.}
\label{fig:ma2_thresh}
\end{center}
\end{figure}

Here we experimented with different hyperparameters for the BNN (BCNN), including different number of bins (classes) in Figure~\ref{fig:ma2_bins}, and different exploitation strategies using the threshold $\delta$ in Figure~\ref{fig:ma2_thresh}. For MA(2), four bins per parameter (16 classes) is the best performing setting, however using five bins (25 classes) results in similar performance. For the exploitation threshold $\delta$, we observe the best overall performance using $\delta=0.05$ when observing log mean EMD. For mean MMD both show similar results. Additionally, a proposed strategy of using an exponentially decaying threshold, i.e., reducing the amount of exploitation per round, performs on a level in between $\delta = 0.05$ and $\delta = 0.0$. 

\subsection{Execution time}
\begin{figure}[H]
\begin{center}
\centerline{\includegraphics[width=\columnwidth]{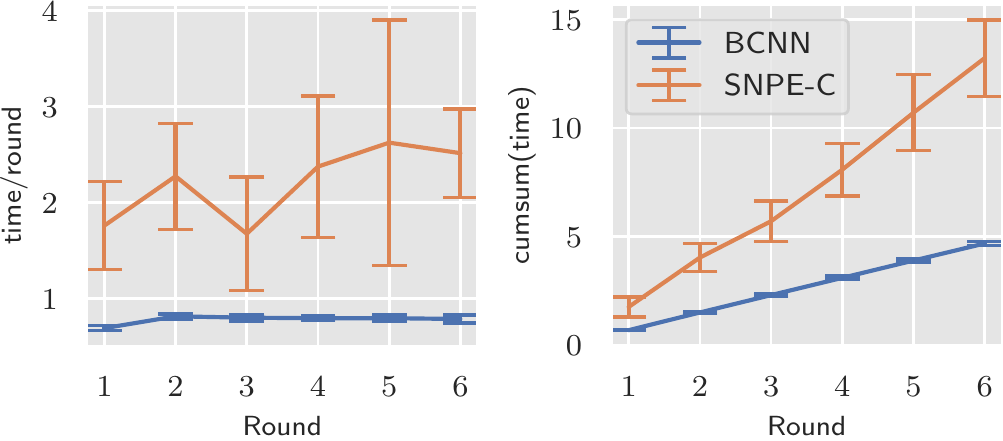}}
\caption{(left) Time (minutes) per round between BCNN and SNPE-C using MAF. (right) Cumulative time (minutes). Both include simulation time per round (3000 simulations/round).  }
\label{fig:ma2_time}
\end{center}
\end{figure}

Figure~\ref{fig:ma2_time} shows the time difference for the inference task between the two implementations used in this study. Note that the BCNN used for MA(2) is one joint model (2D bins), however for the LV model several co-variate BCNN models are trained.

It is important to note that differences in software implementations (TensorFlow versus PyTorch) and approaches to parallelization make direct empirical comparisons challenging. Further, factors such as the early stopping criterion will also have an effect on the result.


\section{Extended Experiments for the Lotka-Volterra (LV) Test Problem:}
The structure of extended experiments for the LV test problem follows the MA(2) setup in the text above. 

\subsection{Qualitative evaluation with additional seeds}
\begin{sidewaysfigure}[ht]
\centering
\includegraphics[width=21cm]{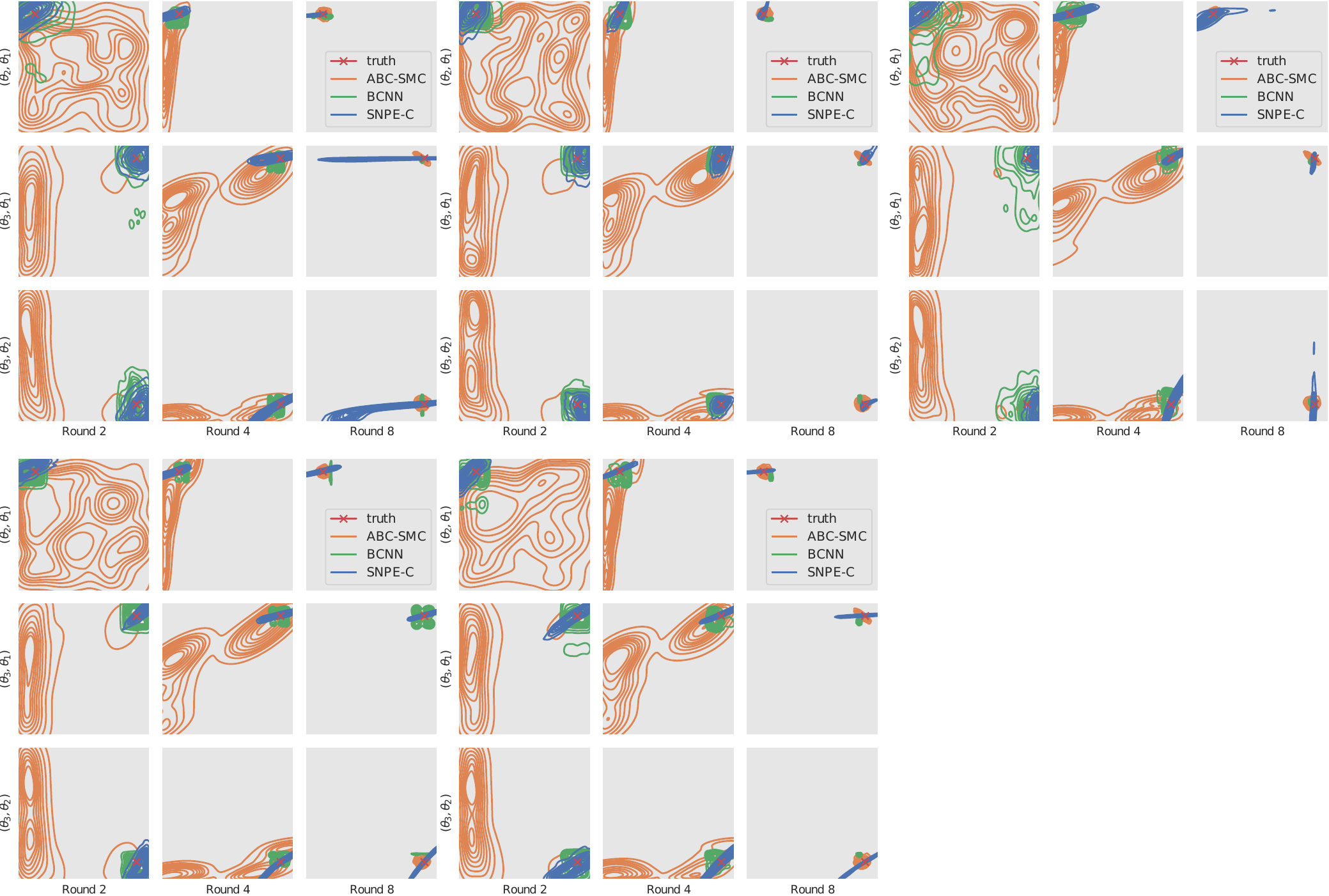}
\caption{Snapshot \# 1-5 of the LV model using ABC-SMC, BCNN, and SNPE-C.}
\label{fig:lv_grid}
\end{sidewaysfigure}

The robustness of the three approaches - BCNN, SNPE-C and ABC-SMC against five different seeds is now investigated for the Lotka-Volterra predator-prey model.
In Figure~\ref{fig:lv_grid}, we show the snapshots and the final two-dimensional marginal distributions for the five independent runs. Comparison of the three approaches shows that the BCNN approach is relatively robust, while the ABC-SMC method converges slowly. It can also be seen that SNPE-C often converges to a different distribution shape as compared to BCNN and ABC-SMC.






\subsection{Embedding Network architectures}
Implementations follow the description in Section \ref{sec:arch_ma2}. The architectures are shown in Table \ref{tab:lv}.
\begin{table}[ht]
\caption{Architecture BCNN (top) and SNPE-C(bottom).}
\label{tab:lv}
\begin{center}
\begin{tabular}{lcccr}
\toprule
Layer & Ouput shape & Weights\\
\midrule
Input    &[(None, 51, 2)]&0\\
Conv1DFlipout (ReLU) &(None, 51, 20)  &260\\
MaxPooling1D    &(None, 10, 20)&0\\
Flatten    & (None, 200)&0\\
DenseFlipout (Softmax)     &(None, 25)&10025\\
\midrule
Total weights &&10285

\end{tabular}

\begin{tabular}{lcccr}
\toprule
Layer & Ouput shape & Weights\\
\midrule
Input    &[(None, 51, 2)]&0\\
Conv1d (ReLU) &(None, 51, 20)  &140\\
MaxPooling1d    &(None, 10, 20)&0\\
view (Flatten)    & (None, 200)&0\\
Linear (ReLU)     &(None, 25)&1608\\
NDE (MAF) &10 hidden, 2 transform&912\\
\midrule
Total weights &&2660

\end{tabular}
\end{center}
\end{table}

\subsection{Hyperparameters of BNN}
\label{sec:hyperLV}
\begin{figure}[H]
\begin{center}
\centerline{\includegraphics[width=\columnwidth]{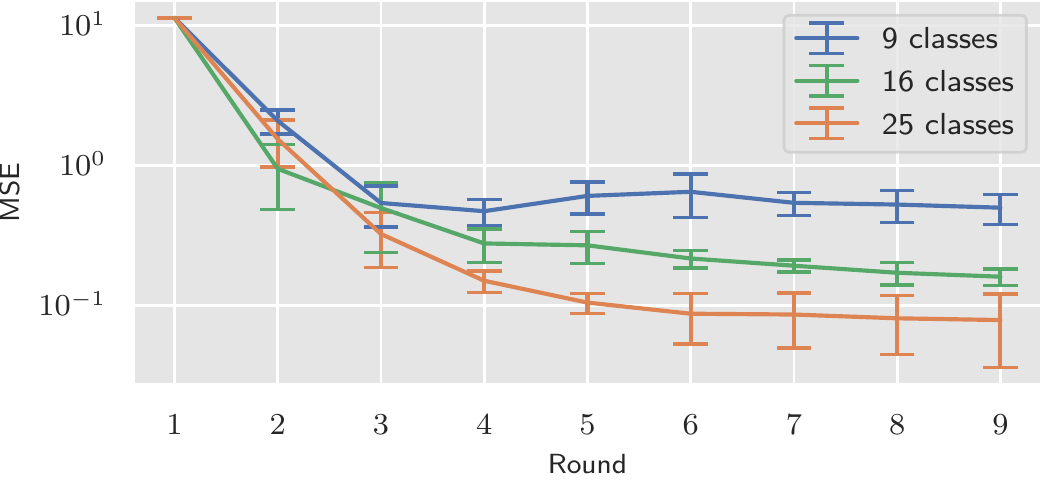}}
\caption{MSE for BCNN with different number of classes corresponding to the square of the number of bins per parameter, on the LV test problem.}
\label{fig:lv_bins}
\end{center}
\end{figure}

As in the case of the MA(2) test problem, the hyperparameters of interest to explore in the context of the BNN aproach for LV include the number of bins (classes) and the extent of exploitation controlled by the threshold $\delta$. In Figure~\ref{fig:lv_bins}, we explore three values for the bins. As the number of bins for the discretization of continuous data increases, so does the convergence speed towards a lower MSE.

\begin{figure}[H]
\begin{center}
\centerline{\includegraphics[width=\columnwidth]{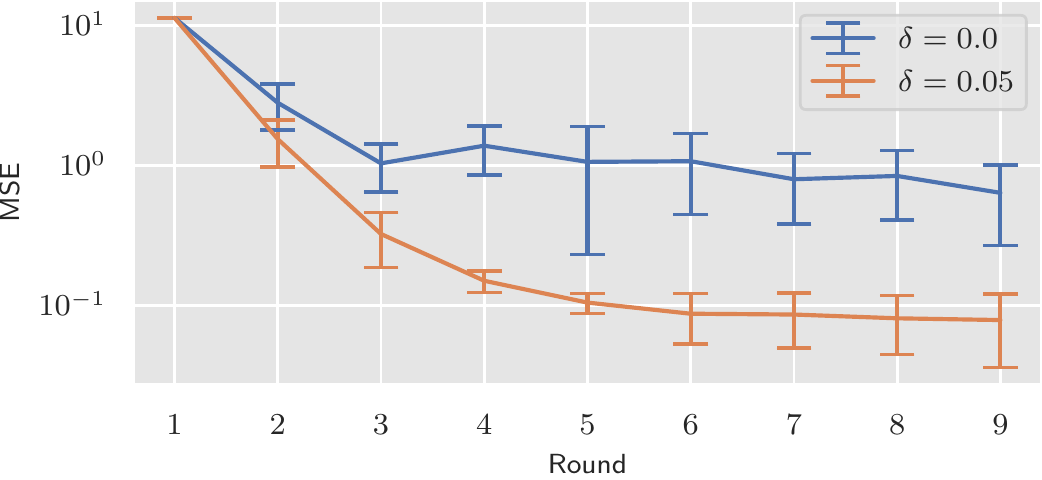}}
\caption{MSE for BCNN when using threshold $\delta$: $0.05$ and $0.0$, on the LV test problem.}
\label{fig:lv_thresh}
\end{center}
\end{figure}

In Figure~\ref{fig:lv_thresh}, we compare the MSE for 8 rounds using a threshold value of $\delta = 0.05$ and $\delta = 0$, i.e., with or without exploitation. It can be seen that incorporating exploitation via $\delta > 0$, results in a more rapid decline in MSE as compared to the case of no exploitation. This is expected, validates the purpose of exploitation in the proposed approach. 

\subsection{Execution time}
\begin{figure}[H]
\begin{center}
\centerline{\includegraphics[width=\columnwidth]{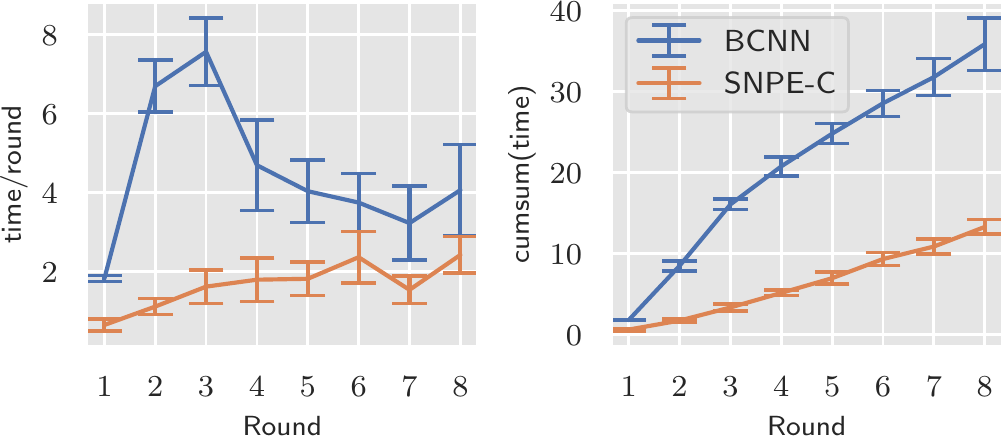}}
\caption{Time per round (minutes) and cumulative time (minutes), for BCNN and SNPE-C, on the LV method.}
\label{fig:lv_time}
\end{center}
\end{figure}

Figure~\ref{fig:lv_time} presents the execution time of the BCNN and SNPE-C approaches. However, we note that several factors affect this comparison. It can be observed that the BCNN approach is slower as compared to SNPE-C owing to the overhead of training times for multiple classifiers.
The number of required classifiers depends upon the number of parameter pair-combinations, e.g., three parameters requires four models, four parameters require six models, etc. The additional training time for the BCNN in case of LV can be reduced via better parallelization of the training process of the three classifiers. Please refer to Figure~\ref{fig:ma2_time} for a comparison of execution times for a single trained model.

\newpage
\onecolumn
\end{appendix}

\end{document}